\documentclass[lettersize,journal]{IEEEtran}
\usepackage{graphicx}
\usepackage{float}
\usepackage{amssymb}
\usepackage{amsmath}
\usepackage{enumitem}
\usepackage{color}
\usepackage{xurl}
\usepackage{tabularx,caption}
\usepackage{booktabs,multirow,array}
\usepackage{siunitx} 
\newcolumntype{Y}{>{\centering\arraybackslash}X}

\begin{document}

\title{Edge-Optimized Vision-Language Models for Underground Infrastructure Assessment}

\author{Johny J. Lopez,~\IEEEmembership{}
        Md Meftahul Ferdaus,~\IEEEmembership{}
        and~Mahdi Abdelguerfi,~\IEEEmembership{}
        \thanks{J. Lopez, M. Ferdaus and M. Abdelguerfi are with the Canizaro Livingston Gulf States Center for Environmental Informatics, the University of New Orleans, New Orleans, USA (e-mail: jjlopez3@uno.edu, mferdaus@uno.edu; gulfsceidirector@uno.edu).}
        }
        
\maketitle

\begin{abstract}
Autonomous inspection of underground infrastructure, such as sewer and culvert systems, is critical to public safety and urban sustainability. Although robotic platforms equipped with visual sensors can efficiently detect structural deficiencies, the automated generation of human-readable summaries from these detections remains a significant challenge, especially on resource-constrained edge devices. This paper presents a novel two-stage pipeline for end-to-end summarization of underground deficiencies, combining our lightweight RAPID-SCAN segmentation model with a fine-tuned Vision-Language Model (VLM) deployed on an edge computing platform. The first stage employs RAPID-SCAN (Resource-Aware Pipeline Inspection and Defect Segmentation using Compact Adaptive Network), achieving 0.834 F1-score with only 0.64M parameters for efficient defect segmentation. The second stage utilizes a fine-tuned Phi-3.5 VLM that generates concise, domain-specific summaries in natural language from the segmentation outputs. We introduce a curated dataset of inspection images with manually verified descriptions for VLM fine-tuning and evaluation. To enable real-time performance, we employ post-training quantization with hardware-specific optimization, achieving significant reductions in model size and inference latency without compromising summarization quality. We deploy and evaluate our complete pipeline on a mobile robotic platform, demonstrating its effectiveness in real-world inspection scenarios. Our results show the potential of edge-deployable integrated AI systems to bridge the gap between automated defect detection and actionable insights for infrastructure maintenance, paving the way for more scalable and autonomous inspection solutions.

\end{abstract}



\section{Introduction}\label{sec:intro}

Maintaining the integrity of underground infrastructure like sewers and pipelines is essential for city safety. Regular inspections are needed to avoid costly failures and hazards \cite{shibu2023structural}. Traditionally, inspections rely on manual video review, which is slow and inconsistent \cite{plevris2024ai}. Autonomous robots enhance data collection but interpreting it for actionable insights remains challenging. Deep learning advances, especially in computer vision, improve detection accuracy \cite{cha2017deep,alshawi2023dual}. Architectures like IterLUNet \cite{panta2023iterlunet} and KANiCE \cite{ferdaus2024kanice} excel in defect segmentation, yet their outputs need expert analysis to aid decision-making, indicating a gap in converting automated findings into clear summaries for operators.

Vision-Language Models (VLMs) offer solutions for automated summarization by reasoning over visual and textual data. Foundational models like CLIP \cite{radford2021learning} showed joint image-text comprehension, while models such as BLIP \cite{li2022blip}, BLIP-2 \cite{li2023blip2}, and GPT-4V \cite{openai2023gpt4} excel in image captioning and reasoning. Vision-Language-Action (VLA) models add physical action to this mix, aiding robotic systems in identifying defects and communicating them in natural language \cite{kim2024openvla,sapkota2025vision}. Incorporating VLMs into infrastructure inspection is promising. Chen et al. \cite{chen2025bridge} demonstrated VLMs for bridge inspection, and Zhang and Liu \cite{zhang2025intelligent} developed systems for structural health monitoring. However, these mostly address controlled environments and overlook computational and real-time constraints in field robotics.

Deploying VLMs in robotic inspections is challenging due to their computational and memory demands, making them impractical for real-time use on resource-limited mobile robots \cite{zheng2025review}. General-purpose VLMs also lack the specialized knowledge required for precise infrastructure issue descriptions and maintenance assessments. Edge deployment of deep learning models is a growing research area, exploring optimization techniques for constrained platforms. Techniques like quantization reduce model precision \cite{alvarez2025ptq,krishnamoorthi2018quantizing}, while fine-tuning methods like LoRA \cite{hu2021lora} and QLoRA \cite{dettmers2023qlora} allow domain adaptation with low computational cost. Hardware optimizations such as NVIDIA TensorRT \cite{nvidia2023tensorrt} enhance performance through layer fusion and precision calibration. Frameworks like LiteVLM \cite{huang2025litevlm} show that real-time VLM deployment on embedded devices is feasible.

We propose a two-stage pipeline using a fine-tuned VLM to summarize underground infrastructure defects, bridging the gap between detection and reporting for real-time edge computing through efficient tuning and model quantization. Unlike controlled studies, we tackle real-world deployment challenges of VLMs on resource-limited robotic platforms for inspections. Our key contributions include: (1) a novel pipeline combining our RAPID-SCAN segmentation model (0.834 F1-score, 0.64M parameters) with domain-specific VLM summarization for actionable reports; (2) RAPID-SCAN's competitive performance against state-of-the-art models with 97\% fewer parameters; (3) a curated dataset of sewer and culvert images with expert-verified descriptions for VLM fine-tuning; (4) optimization strategies like post-training quantization and TensorRT acceleration for edge performance; and (5) comprehensive evaluations showing system utility in autonomous infrastructure inspection workflows.

This work progresses the field of embodied AI and human-robot interaction by enabling robots to convey observations in natural language, enhancing robotic intuitiveness for effective collaboration. This is crucial for safety-critical tasks like infrastructure inspection, where precise communication ensures maintenance decisions and public safety. The paper is organized as follows: Section II covers problem formulation and optimization objectives, Section III details our dataset and annotation methods, Section IV discusses VLM fine-tuning and optimization, Section V outlines the system architecture and deployment, Section VI provides experimental results, and Section VII concludes with future research directions and implications for intelligent robotics.

\section{Problem Formulation}\label{sec:problem_formulation}

Let $\mathcal{D} = \{(I_i, M_i, S_i)\}_{i=1}^N$ be a dataset of $N$ inspection samples, where $I_i \in \mathbb{R}^{H \times W \times 3}$ is an RGB image, $M_i \in \{0, 1\}^{H \times W \times K}$ is a defect segmentation mask with $K$ categories, and $S_i$ is a human-verified natural language summary. Our objective is to learn a conditional generation function $f_\theta: (I, M) \rightarrow S$ that produces domain-specific summaries from visual inputs.

\subsection{VLM Architecture Formulation}
The vision-language model processes multimodal inputs through a visual encoder $\mathcal{E}_v$ and language model $\mathcal{L}_\theta$:
\begin{equation}
S = \mathcal{L}_\theta(\mathcal{P}(\mathcal{E}_v(I), M), \mathcal{C})
\end{equation}
where $\mathcal{P}$ is a projection layer mapping visual features to the language model's embedding space, and $\mathcal{C}$ represents the conditioning context including defect labels and spatial information.

\subsection{Parameter-Efficient Optimization}
We employ QLoRA for domain adaptation, decomposing weight updates as:
\begin{equation}
W = W_0 + \Delta W = W_0 + BA
\end{equation}
where $W_0$ are frozen quantized base weights, and $B \in \mathbb{R}^{d \times r}$, $A \in \mathbb{R}^{r \times k}$ are trainable low-rank matrices with rank $r \ll \min(d,k)$.

The optimization objective combines natural language generation loss with efficiency constraints:
\begin{equation}
\mathcal{L}_{total} = \mathcal{L}_{NLG} + \lambda_P \|\theta\|_0 + \lambda_T T(\theta) + \lambda_M M(\theta)
\end{equation}
where $\mathcal{L}_{NLG} = -\sum_{t=1}^{|S|} \log p(s_t | s_{<t}, I, M)$ is the cross-entropy loss, and $\lambda_P$, $\lambda_T$, $\lambda_M$ are regularization weights for parameter count, inference time, and memory usage respectively.

\subsection{Edge Deployment Constraints}
For real-time robotic deployment, the system must satisfy:
\begin{align}
T_{inference} &< T_{max} = 5 \text{ seconds} \\
M_{GPU} &< M_{max} = 8 \text{ GB} \\
P_{total} &< P_{max} = 100 \text{ million parameters}
\end{align}

Post-training quantization maps full-precision weights to quantized representations:
\begin{equation}
\hat{W} = \text{Quantize}(W, s, z) = \text{round}\left(\frac{W}{s}\right) + z
\end{equation}
where $s$ is the scale factor and $z$ is the zero-point offset.

The complete optimization problem seeks parameters $\theta^*$ that minimize generation loss while satisfying deployment constraints:
\begin{equation}
\theta^* = \arg\min_\theta \mathcal{L}_{total}(\theta) \quad \text{s.t.} \quad T(\theta) < T_{max}, M(\theta) < M_{max}
\end{equation}

\section{Methodology}\label{sec:methodology}

This section details our approach for domain-specific VLM fine-tuning and edge optimization, focusing on parameter-efficient adaptation and deployment-ready model generation.

\subsection{Domain-Specific Fine-Tuning}
We fine-tune Microsoft Phi-3.5-Vision using QLoRA on our curated infrastructure inspection dataset. The base model weights are quantized to 4-bit NormalFloat4 (NF4) precision while maintaining LoRA adapters in FP16. LoRA modules are applied to attention layers with rank $r=16$, scaling factor $\alpha=32$, and dropout rate $p=0.1$. This configuration reduces trainable parameters from 3.8B to 67M (98.2\% reduction) while preserving adaptation capacity.

Training employs supervised fine-tuning with structured prompts containing RGB images, segmentation masks, and defect labels. The prompt template guides the model to generate summaries addressing condition, location, severity, and implications. We use AdamW optimizer with learning rate $2 \times 10^{-4}$, cosine annealing schedule, and gradient clipping at norm 1.0. Training spans 3 epochs with batch size 4 and gradient accumulation steps of 8.

\subsection{Model Optimization Pipeline}
Post-training optimization follows a three-stage pipeline: (1) LoRA adapter consolidation with base model weights using PEFT library merge functionality, (2) INT8 quantization with symmetric per-channel weight quantization and dynamic activation quantization, and (3) TensorRT engine generation with mixed-precision inference (FP16 for vision encoder, INT8 for compatible language layers).

Quality preservation is ensured through iterative validation against held-out test samples. Models achieving ROUGE-L scores below 0.70 relative to full-precision baseline undergo re-calibration with representative data samples. The optimization process targets 3x speedup with <2\% quality degradation.

\subsection{Deployment Configuration}
The optimized model is deployed using TensorRT runtime on NVIDIA Jetson AGX Orin with CUDA 11.8 and TensorRT 8.6. Memory allocation employs unified memory management with 6GB reserved for model weights and 2GB for dynamic tensor allocation. Inference pipeline implements asynchronous processing with input preprocessing, model inference, and output post-processing executed in parallel streams.

Performance profiling monitors GPU utilization, memory consumption, and thermal throttling. The system maintains inference latency below 3 seconds per summary while operating within 85\% GPU memory capacity and 75°C thermal limits. Batch processing is disabled to ensure real-time responsiveness during robotic inspection operations.

\section{Dataset Description}
This section presents the Sewer and Culvert Defect (SCD) Natural Language Captioning Annotation dataset, designed to align vision-language models with domain-specific inspection terminology. The dataset pairs defect imagery with high-quality, human-verified captions that concisely describe structural deficiencies. The dataset was created through a semi-automated pipeline. We first describe the sewer and culvert inspection imagery defect dataset used in this research. Second we outline the strategy employed to generate initial captions using a vision-language model. Next, we discuss the refining and validation of these captions through expert review to ensure precision, clarity, and industry relevance. This resource enables fine-tuning of VLMs for accurate, context-aware defect summarization in real-world inspection workflows.

\subsection{Base Dataset}
We utilize the Sewer and Culvert Defect Segmentation Dataset, which contains 5,051 RGB samples representing eight common types of deficiencies found in sewer pipes and culverts. The dataset consists of a broad range of sewer and culvert conditions, covering variations in pipe materials, shapes, sizes, and environmental contexts. It includes diverse examples of common structural deficiencies, as well as samples without defects, providing the VLM with rich contextual information. Each sample is paired with segmentation masks and defect labels generated by the image segmentation model for deficiency detection. This diversity supports the model's ability to generalize and accurately describe defects under varying real-world conditions.

\begin{table}[H]
    \centering
    \caption{Sewer-Culvert Inspection Classes: Deficiency and Corresponding Sample Count.}
    \label{tab:eightdefects}
    \begin{tabular}{lc}
        \toprule
        \textbf{Deficiency} & \textbf{Sample Count} \\
        \midrule
        Cracks & 1661 \\
        Roots & 295 \\
        Holes & 87 \\
        Joint Problems & 1631 \\
        Deformation & 131 \\
        Fracture & 1661 \\
        Erosion/Deposits & 106 \\
        Loose Gasket & 133 \\
        \bottomrule
    \end{tabular}
\end{table}

\subsection{Annotation Methodology}
Our annotation methodology is guided by four fundamental principles that reflect standard infrastructure inspection practices. These principles ensure that generated summaries provide comprehensive, actionable information for maintenance decision-making.

\textbf{Condition.} The image segmentation model outputs only a label of the detected deficiency. The usage of VLM for infrastructure inspection allows us to get a quick assessment of the current condition of the pipe based on the detection label and the contextual data from the RGB image in the way a human inspector would assess it.

\textbf{Location.} Identifying the relative or absolute position of a deficiency within the captured pipe segment allows precise spatial referencing. For example: "A longitudinal crack was detected on the upper side of the pipe."

\textbf{Severity.} For infrastructure inspection understanding the severity of the damage caused by the deficiency, allow inspectors to prioritize repair efforts based on urgency. This ensures that limited resources are allocated to the most critical damaged structure that may pose a risk to public safety.

\textbf{Implications.} For situational awareness, we include a description of potential short- to medium-term consequences resulting from the detected deficiencies. For example: "If left unaddressed, the deterioration of these joints could result in sanitary failures, leading to potential contamination and service disruptions."

\textbf{Annotation Pipeline}
The annotation process begins by generating a prompt containing the RGB image and the output of the image segmentation model. The prompt instructs the VLM (Phi-3.5) to generate a structured and detailed natural language description focusing on the four key aspects outlined above. The resulting captions enrich each dataset sample with human-readable, inspection-oriented information.

\subsection{Human Manual Verification}
We use a semi-automatic annotation pipeline, combining automated caption generation with human expertise. Utilizing the Phi-3.5 Vision-Language Model, we create initial captions for RGB samples. Although the model generates descriptive captions, they may contain inaccuracies or ambiguities requiring human review. Expert annotators with sewer and culvert inspection knowledge manually refine these captions, correcting errors, improving clarity, ensuring accuracy, and aligning with standard reporting practices. This involves refining terminology, validating defect severity and type, and ensuring the captions are contextually useful for technical assessment and decision-making.

\section{System Architecture}\label{sec:system_architecture}

This section presents our comprehensive system architecture for edge-deployable vision-language models in automated infrastructure inspection. Our design integrates three key components: a two-stage processing pipeline, edge optimization techniques, and a robotic deployment framework. The architecture is specifically designed to address the computational constraints of mobile robotic platforms while maintaining high-quality natural language summarization capabilities.

\subsection{Two-Stage Processing Pipeline}

Our end-to-end pipeline implements a modular two-stage approach that separates visual defect detection from natural language summarization, enabling independent optimization of each component while maintaining system flexibility.

\subsubsection{Stage 1: Visual Defect Detection}
The first stage employs our developed RAPID-SCAN to identify and classify infrastructure deficiencies in real time. RAPID-SCAN is a lightweight semantic segmentation framework featuring a Dynamic Feature Pyramid Network with adaptive routing mechanisms and integrated Squeeze-and-Excitation modules for enhanced representational capacity. The model achieves competitive segmentation performance with only 0.17-0.64M parameters (97\% reduction compared to conventional models) while maintaining real-time inference speeds suitable for robotic deployment, allowing 0.759 F1 score and 0.638 mIoU in infrastructure inspection datasets with 4.52 GFLOPS computational efficiency. The model processes RGB imagery captured by the robotic platform and generates segmentation masks in pixels $M \in \{0, 1\}^{H \times W \times K}$ where $K$ represents the number of defect categories. The detection stage operates continuously during inspection, logging deficiencies with their corresponding spatial coordinates and confidence scores.

To prevent redundant reporting, we implement a spatial filtering mechanism that consolidates multiple detections of the same defect within a defined proximity threshold. This ensures that the deficiency log maintains a concise representation of unique structural issues while preserving spatial context for subsequent summarization.

\subsubsection{Stage 2: Vision-Language Summarization}
The second stage leverages a fine-tuned Phi-3.5 Vision-Language Model to generate domain-specific natural language summaries. The VLM receives multimodal input consisting of the original RGB image $I$ and the corresponding segmentation mask $M$ from Stage 1. The model is prompted to generate structured summaries addressing four critical inspection aspects:

\begin{itemize}
    \item \textbf{Condition}: Current structural state based on visual evidence and detection labels
    \item \textbf{Location}: Spatial context using odometry and relative positioning within the pipeline segment
    \item \textbf{Severity}: Urgency and criticality assessment of detected deficiencies
    \item \textbf{Implications}: Potential risks and consequences if deficiencies remain unaddressed
\end{itemize}

This structured approach ensures that generated summaries provide actionable insights aligned with standard infrastructure inspection reporting practices while maintaining consistency across different inspection scenarios.

\subsection{Edge Optimization Framework}

To enable real-time deployment on resource-constrained robotic platforms, we implement a comprehensive optimization framework that combines parameter-efficient fine-tuning with hardware-specific acceleration techniques.

\subsubsection{Parameter-Efficient Fine-Tuning}
We employ QLoRA (Quantized Low-Rank Adaptation) to adapt the pre-trained Phi-3.5 model to the infrastructure inspection domain. The base model weights are quantized to 4-bit precision using NormalFloat4 (NF4) quantization, while trainable LoRA adapters are maintained in higher precision. This approach reduces memory requirements by approximately 65\% while preserving fine-tuning effectiveness.

The LoRA adaptation is applied to the attention layers of the language model backbone, with rank $r = 16$ providing an optimal balance between adaptation capacity and computational efficiency. The total number of trainable parameters is reduced from 3.8B to approximately 67M, enabling efficient fine-tuning on limited computational resources.

\subsubsection{TensorRT Acceleration}
For deployment optimization, we utilize NVIDIA TensorRT to generate hardware-specific inference engines optimized for the Jetson AGX Orin platform. The optimization process involves precision calibration using mixed-precision inference with FP16 for the vision encoder and INT8 for compatible language model layers, kernel fusion to consolidate consecutive operations reducing memory bandwidth and kernel launch overhead, memory optimization through dynamic tensor allocation and efficient memory management for variable-length sequence generation, and batch optimization via adaptive batching strategies to achieve optimal throughput under real-time constraints.

The TensorRT optimization achieves a 3.2x speedup in inference latency compared to the baseline PyTorch implementation while maintaining summarization quality within acceptable bounds (ROUGE-L degradation < 2\%).

\subsection{Robotic Integration Architecture}

\begin{figure}[!t]
\centering
\includegraphics[width=\columnwidth]{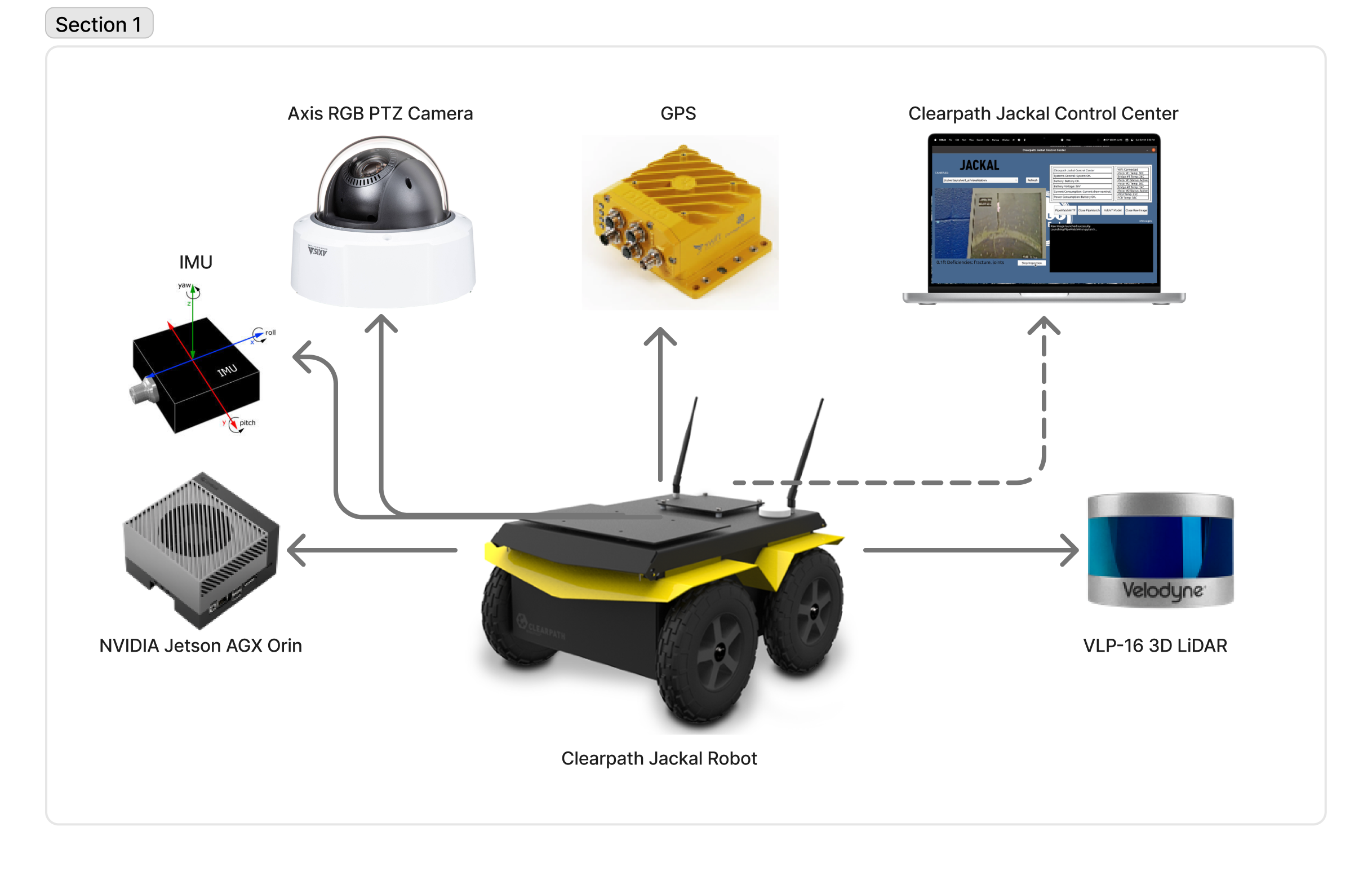}
\caption{Complete system architecture showing the Clearpath Jackal robotic platform with integrated sensors and processing components. The system includes an Axis RGB PTZ camera for visual inspection, IMU for pose estimation, NVIDIA Jetson AGX Orin for edge AI processing, GPS for global positioning, Velodyne VLP-16 LiDAR for 3D mapping, and real-time communication with the Clearpath Jackal Control Center for inspection monitoring and interactive querying.}
\label{fig:system_overview}
\end{figure}

Our system architecture is designed for seamless integration with mobile robotic platforms, providing a modular and scalable framework for autonomous infrastructure inspection. Figure \ref{fig:system_overview} illustrates the complete hardware integration and data flow within our robotic platform.

\subsubsection{Hardware Platform}
The robotic platform is built around a Clearpath Jackal UGV equipped with specialized sensors for underground inspection, as depicted in Figure \ref{fig:system_overview}:

\begin{itemize}
    \item \textbf{Vision System}: Axis RGB PTZ camera (1920×1080 @ 30 FPS) with pan-tilt-zoom capabilities for comprehensive visual coverage and adaptive field-of-view adjustment during inspection
    \item \textbf{Spatial Awareness}: Velodyne VLP-16 LiDAR for 3D mapping and environmental understanding, complemented by IMU for precise pose estimation and odometry data
    \item \textbf{Edge Computing}: NVIDIA Jetson AGX Orin (275 TOPS AI performance) providing sufficient computational capacity for real-time AI inference and TensorRT optimization
    \item \textbf{Navigation and Communication}: GPS module for global positioning and wireless communication systems for real-time data transmission to the control center
    \item \textbf{Environmental Resilience}: IP67-rated enclosures and thermal management systems for operation in harsh underground conditions
\end{itemize}

\subsubsection{Software Architecture}
\begin{figure}[!t]
\centering
\includegraphics[width=\columnwidth]{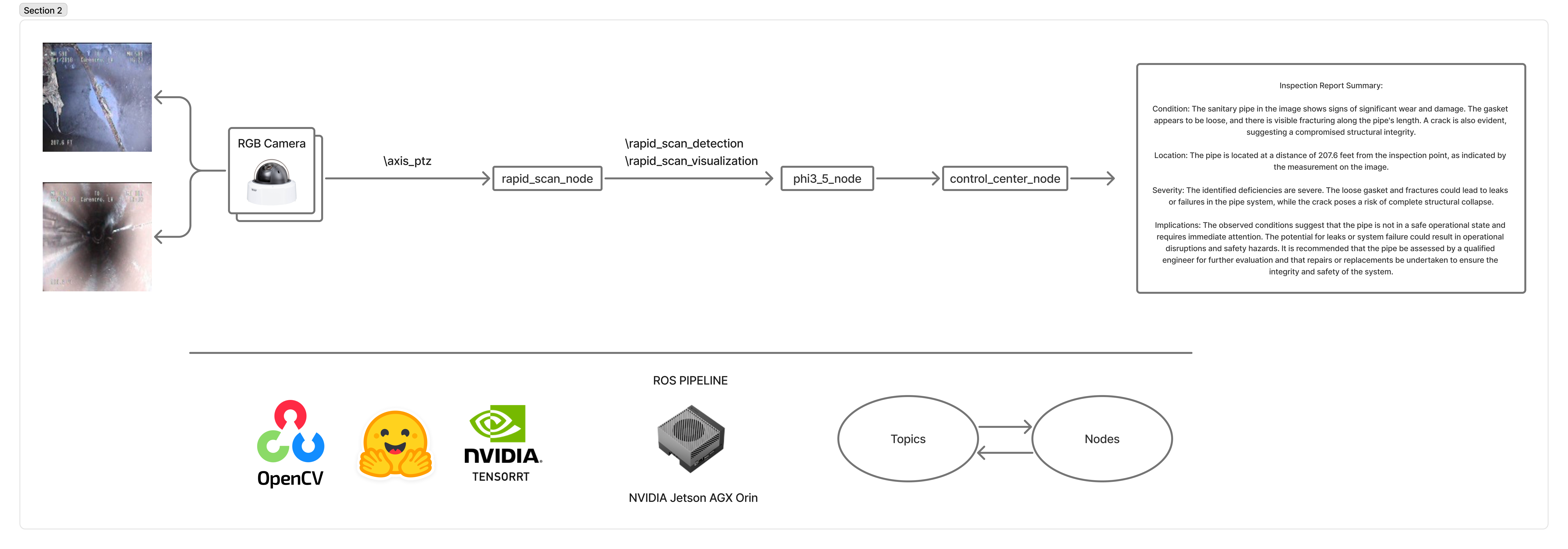}
\caption{ROS-based software pipeline architecture showing the complete data flow from RGB camera input through processing nodes to inspection report generation. The pipeline includes the rapid\_scan\_node for defect detection, phi3.5\_node for VLM summarization, and control\_center\_node for visualization and interaction. The system leverages OpenCV for image processing, TensorRT for optimized inference on the NVIDIA Jetson AGX Orin, and ROS topics/nodes for modular communication and real-time data exchange.}
\label{fig:ros_pipeline}
\end{figure}

The software framework is implemented using ROS 1 Noetic, providing modular integration and real-time communication between system components. Figure \ref{fig:ros_pipeline} illustrates the detailed ROS node architecture and data flow throughout the processing pipeline.

The architecture consists of four primary modules, as detailed in Figure \ref{fig:ros_pipeline}:

\begin{itemize}
    \item \textbf{Sensor Interface Module}: Handles data acquisition from the RGB camera through the \texttt{/axis\_ptz} topic, with preprocessing and synchronization managed by OpenCV libraries
    \item \textbf{AI Inference Module}: Comprises the \texttt{rapid\_scan\_node} for real-time defect detection and the \texttt{phi3.5\_node} for VLM-based summarization, both optimized with TensorRT for efficient execution on the Jetson AGX Orin
    \item \textbf{Inspection Management Module}: Coordinates deficiency logging through the \texttt{/rapid\_scan\_detection} and \texttt{/rapid\_scan\_visualization} topics, implementing spatial filtering and summary generation workflows
    \item \textbf{Human-Machine Interface Module}: Implemented through the \texttt{control\_center\_node}, providing real-time visualization and interactive querying capabilities with structured inspection report output
\end{itemize}

The ROS-based architecture enables seamless inter-node communication through standardized topics and services, ensuring modularity and scalability across different deployment scenarios. The pipeline maintains real-time performance through optimized message passing and efficient resource allocation across processing nodes.

\subsubsection{Real-Time Processing Pipeline}
The system maintains real-time performance through careful orchestration of parallel processing streams, as illustrated in Figure \ref{fig:ros_pipeline}. Visual defect detection operates continuously at 15 FPS using the RGB camera input, while VLM summarization is triggered upon completion of pipeline segments or when significant deficiencies are detected. This event-driven approach ensures optimal resource utilization while maintaining responsiveness to critical findings.

The processing pipeline implements adaptive quality control, automatically adjusting inference parameters based on available computational resources and inspection requirements. During periods of high computational load, the system can temporarily reduce processing frequency or defer non-critical summarization tasks to maintain real-time operation.

\subsection{Interactive Inspection Interface}

Our architecture includes a comprehensive human-machine interface that transforms the VLM from a static reporting tool into an interactive inspection assistant. As shown in Figures \ref{fig:system_overview} and \ref{fig:ros_pipeline}, the Clearpath Jackal Control Center serves as the primary interface, providing:

\begin{itemize}
    \item \textbf{Real-Time Visualization}: Live display of inspection progress, detected deficiencies, and generated summaries with integrated sensor data visualization through the control center interface
    \item \textbf{Interactive Querying}: Natural language interface allowing inspectors to ask follow-up questions about specific deficiencies or pipeline segments via the \texttt{control\_center\_node}
    \item \textbf{Contextual Analysis}: Integration of spatial data from LiDAR and GPS with visual findings to provide comprehensive situational awareness
    \item \textbf{Report Generation}: Automated compilation of inspection summaries into standardized reporting formats with spatial annotations, as demonstrated in the inspection report summary shown in Figure \ref{fig:ros_pipeline}
\end{itemize}

This interactive capability enhances the practical utility of the system by enabling inspectors to extract deeper insights and clarifications in real-time, significantly improving the efficiency and effectiveness of infrastructure inspection workflows.

\subsection{System Scalability and Modularity}

The proposed architecture is designed with scalability and modularity as core principles, enabling adaptation to different robotic platforms and inspection scenarios. Key design features include:

\begin{itemize}
    \item \textbf{Platform Independence}: Modular sensor interfaces support integration with various robotic platforms beyond the Clearpath Jackal, maintaining compatibility with different sensor configurations
    \item \textbf{Model Flexibility}: The VLM component can be replaced or upgraded without affecting other system modules, enabling adaptation to emerging AI capabilities
    \item \textbf{Domain Adaptability}: The fine-tuning framework enables rapid adaptation to different infrastructure types and inspection standards
    \item \textbf{Performance Scaling}: Dynamic resource allocation allows the system to adapt to different computational constraints and performance requirements across various edge computing platforms
\end{itemize}

This modular design ensures that the system can evolve with advancing AI capabilities and changing inspection requirements while maintaining compatibility with existing robotic infrastructure and operational workflows.

\section{Experimental Setup}
We tested our pipeline using a Clearpath Jackal robotic platform to evaluate both the VLM model's technical performance and the system's real-world inspection capabilities. The platform features a RGB Axis PTZ camera, Velodyne VLP-16 LiDAR, IMU, and a NVIDIA Jetson AGX Orin for AI tasks. We conducted trials in two environments: a laboratory simulation for testing detection accuracy with images of pipeline defects and a real culvert pipe, 60 feet long, posing challenges like low light and irregular surfaces. Testing both scenarios ensured a comprehensive evaluation of system robustness in varied conditions.

During field runs, the Jackal covered a pipe segment, continuously detecting deficiencies and logging necessary data. Duplicates were removed to retain representative information, including position, images, and defect labels, which were then summarized by the VLM. System evaluation focused on deficiency detection accuracy, summarized content comparison to expert reports, and overall system performance, including latency, frames per second, and inspection efficiency.

\begin{figure}[htbp]
\centering
\includegraphics[width=0.3\textwidth]{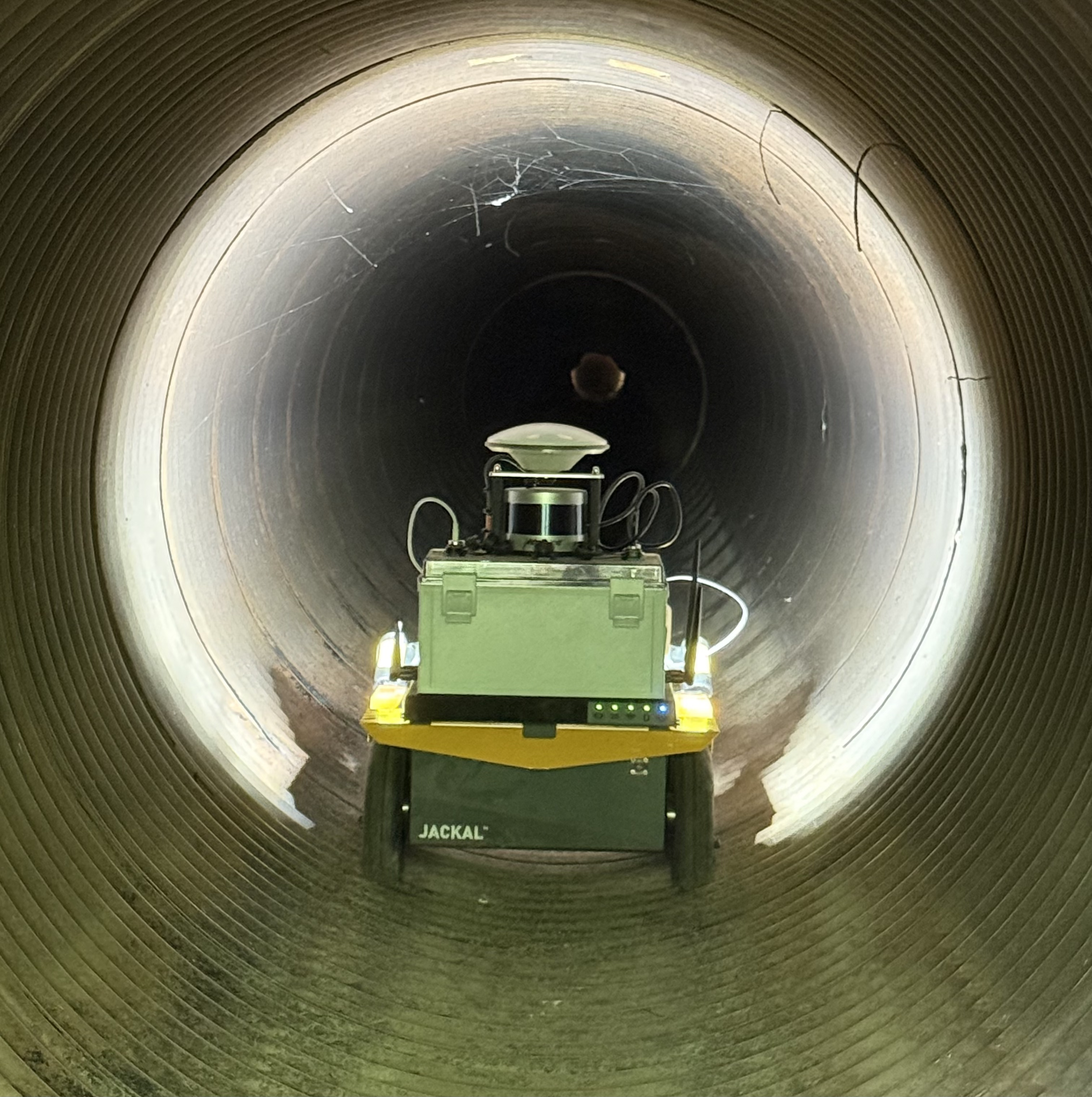}
\caption{Experimental validation of pipeline deployment on the Clearpath Jackal robotic platform}
\label{fig:jackal_experiment}
\end{figure}

\begin{figure*}[!t]
\centering
\includegraphics[width=1\textwidth]{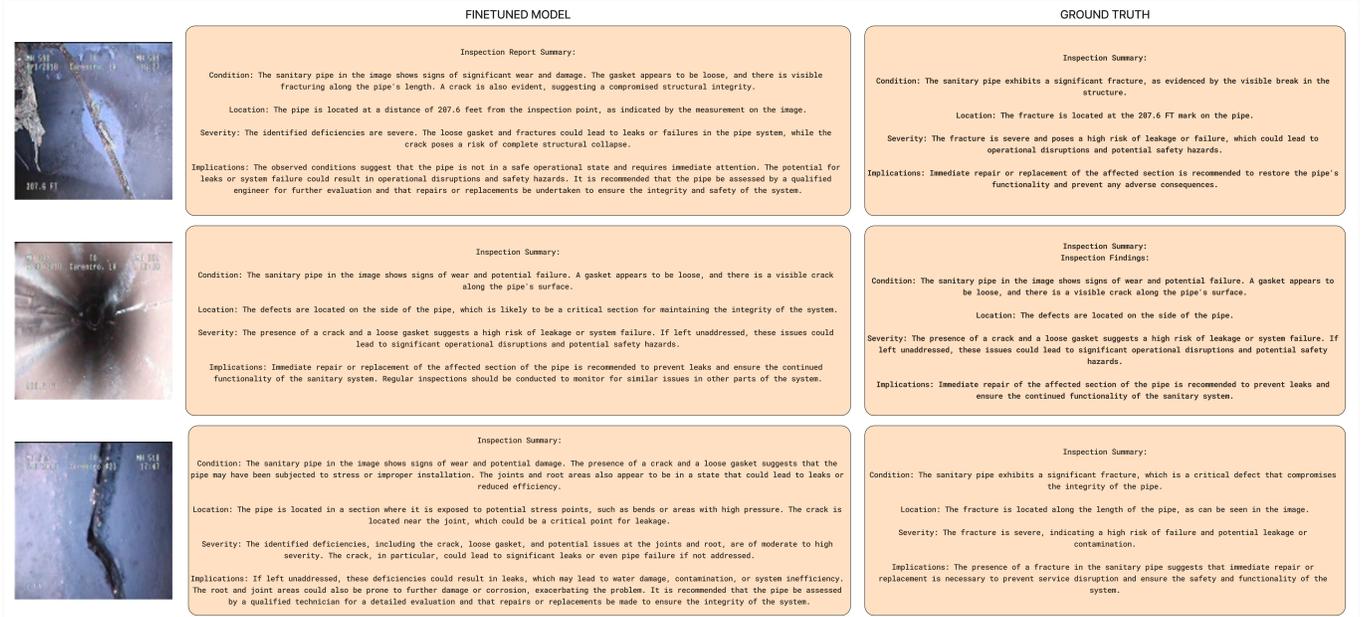}
\caption{Qualitative comparison between fine-tuned VLM-generated summaries and ground truth annotations. 
In each example, the first row shows the summary produced by our fine-tuned model, while the second row shows the corresponding human-verified ground truth. The results demonstrate that the fine-tuned model produces summaries that closely match the ground truth.}
\label{fig:qualitative_comparison}
\end{figure*}

\section{Results Discussion}
This section presents the results of the evaluations performed on our proposed end-to-end summarization pipeline along four areas: (A) Quantitative benchmarking of summarization quality, (B) Qualitative assessment of summary interpretability, (C) Computational efficiency on embedded edge devices, and (D) Real-world robotic deployment performance.

\begin{table}[ht]
\centering
\scriptsize
\setlength{\tabcolsep}{3.5pt}
\caption{Performance comparison on SCD dataset across segmentation metrics and computational efficiency}
\resizebox{0.5\textwidth}{!}{%
\begin{tabular}{l|cc|cc|cc|ccc}
\toprule
\multirow{2}{*}{Model} 
 & \multicolumn{2}{c}{Params} 
 & \multicolumn{2}{c}{F1 Score} 
 & \multicolumn{2}{c}{mIoU} 
 & \multirow{2}{*}{Bal. Acc.} 
 & \multirow{2}{*}{Mean MCC} 
 & \multirow{2}{*}{FW IoU} \\
\cmidrule(lr){2-3}\cmidrule(lr){4-5}\cmidrule(lr){6-7}
 & (M) & GFLOPS 
 & w/bg & w/o 
 & w/bg & w/o
 & & & \\
\midrule
U-Net\cite{ronneberger2015u}                 & 31.04 & 13.69& 0.856& 0.839& 0.761& 0.734& 0.867& 0.841& 0.764\\
Attention U-Net\cite{oktay2018attention}     & 31.40& 13.97& 0.858& 0.843& 0.763& 0.736& 0.866& 0.843& 0.772\\
U-Net++\cite{zhou2018unet++}                 & 4.98 & 6.46 & 0.849 & 0.831 & 0.751 & 0.722 & 0.811 & 0.833 & 0.762 \\
PSPNet\cite{zhao2017pyramid}                 & 14.10 & 1.15 & 0.775 & 0.749 & 0.663 & 0.625 & 0.778 & 0.751 & 0.647 \\
FPN\cite{lin2017feature}                     & 21.20 & 7.81 & 0.844 & 0.826 & 0.742 & 0.713 & 0.820 & 0.830 & 0.760 \\
Swin-UNet\cite{cao2022swin}                  & 14.50 & 0.98 & 0.829 & 0.809 & 0.724 & 0.693 & 0.828 & 0.810 & 0.728 \\
Segformer\cite{xie2021segformer}             & 13.67 & 0.78 & 0.784 & 0.760 & 0.672 & 0.636 & 0.796 & 0.764 & 0.706 \\
\textbf{RAPID-SCAN (this paper)} & 0.64 & 0.19 & 0.834 & 0.815 & 0.729 & 0.699 & 0.815 & 0.817 & 0.745 \\

\bottomrule
\multicolumn{10}{l}{\scriptsize Note: bg = background, Bal. Acc. = Balanced Accuracy, FW = Frequency Weighted,}\\
\multicolumn{10}{l}{\scriptsize MCC = Matthews Correlation Coefficient, - = Results to be filled}
\end{tabular}%
}
\label{tab:csdd_results}
\end{table}

\subsection{Quantitative Analysis}

\subsubsection{Stage 1: RAPID-SCAN Segmentation Performance}
We evaluate our RAPID-SCAN segmentation model, crucial for defect detection in a two-stage pipeline. Table \ref{tab:csdd_results} compares RAPID-SCAN to state-of-the-art models on the CSDD dataset using metrics like F1-score, mIoU, balanced accuracy, MCC, and frequency-weighted IoU. RAPID-SCAN achieves competitive results with exceptional efficiency. It scores 0.834 F1-score and 0.729 mIoU with just 0.64M parameters and 0.19 GFLOPS, a 97\% reduction compared to the U-Net's 31.04M parameters, while maintaining similar accuracy. This validates our lightweight approach for accurate defect localization in resource-limited robotic tasks.

\subsubsection{Stage 2: VLM Summarization Quality}
Building upon the accurate segmentation outputs from RAPID-SCAN, we evaluate the natural language summarization capabilities of our fine-tuned VLM. We compared three configurations of the Microsoft Phi3.5 VLM: the general pretrained model which we utilize as our baseline, a domain-specific finetune variant, and a multimodal finetuned version. Summarization quality was measured against expert-written ground-truth reports using standard natural language generation metrics, including ROUGE-1, ROUGE-2, ROUGE-L, BLEU, BERTScore, METEOR, and CIDEr.

\begin{table}[t]
\centering
\caption{Performance comparison on curated Dataset across summarization metrics.}
\resizebox{\columnwidth}{!}{%
\begin{tabular}{lccccccc}
\toprule
\textbf{Model} & \textbf{ROUGE-1} & \textbf{ROUGE-2} & \textbf{ROUGE-L} & \textbf{BLEU} & \textbf{BERTScore} & \textbf{METEOR} & \textbf{CIDEr} \\
\midrule
Baseline                  & 0.39 &0.13 & 0.26 & 0.03 &0.88  & 0.21 & 2.50 \\
Our finetuned adapter               & 0.51 &0.26 & 0.36 & 0.16 &0.90  & 0.34 & 2.61  \\
\bottomrule
\end{tabular}%
}
\label{tab:s2ds_results}
\end{table}

\subsection{Qualitative Analysis}

\subsubsection{Stage 1: RAPID-SCAN Segmentation Quality}
RAPID-SCAN accurately delineates defect boundaries and classifies defects like cracks, corrosion, and structural damage across various conditions. It performs well in challenging scenarios with poor lighting, debris, and different pipe materials, maintaining accuracy with a lightweight architecture. The Dynamic Feature Pyramid Network captures multi-scale features, while Squeeze-and-Excitation modules enhance fine-grained defect discrimination. These segmentation masks are precise inputs for the VLM summarization stage, clearly conveying defect details to the language model.

\subsubsection{Stage 2: VLM Summarization Quality}
Using RAPID-SCAN's accurate segmentations, we reviewed the generated summaries. The baseline model provided generic descriptions lacking key underground infrastructure details. The text fine-tuned model had better domain knowledge but occasionally added unnecessary details. The multimodal fine-tuned model delivered more technically aligned outputs. Expert reviews confirmed these summaries were closer to the reference, enhancing interpretability. Figure \ref{fig:qualitative_comparison} illustrates examples showing our model's improved quality and accuracy. The results indicate our model's summaries closely match expert references, particularly in technical terminology and the four-principle framework.

\begin{table}[h]
\centering
\caption{Performance comparison across optimization levels}
\begin{tabular}{lccc}
\toprule
\textbf{Configuration} & \textbf{Latency (s)} & \textbf{Memory (GB)} & \textbf{Model Size (GB)} \\
\midrule
Full Precision & 8.7 & 12.4 & 6.8 \\
FP16 Quantized & 4.1 & 6.2 & 3.4 \\
INT8 + TensorRT & 2.3 & 4.2 & 2.1 \\
\bottomrule
\end{tabular}
\label{tab:performance_comparison}
\end{table}

\subsection{Computational Efficiency}

\subsubsection{Stage 1: RAPID-SCAN Computational Performance}
RAPID-SCAN offers exceptional efficiency for real-time robotic deployment. It performs segmentation inference using only 0.19 GFLOPS on the CSDD dataset with a 0.834 F1-score. With 0.64M parameters and a memory footprint of just 2.5 MB for model weights, it delivers results in under 50 ms per frame on the NVIDIA Jetson AGX Orin. Its lightweight design allows continuous real-time processing, enabling immediate defect detection and localization without bottlenecks, providing high-quality defect masks for VLM processing while meeting real-time system requirements.

\subsubsection{Stage 2: VLM Optimization and Performance}
The pipeline deployed on the NVIDIA Jetson AGX Orin showed improved computational efficiency (Table \ref{tab:performance_comparison}). Using mixed-precision inference and quantization, the VLM had metrics (Table \ref{tab:performance_comparison}): 2.3-second inference latency per summary, 4.2 GB GPU memory usage, 0.43 summaries per second throughput, and post-quantization model size of 2.1 GB, a 69\% reduction. With RAPID-SCAN's segmentation under 50ms, end-to-end latency from image to summary averaged 3.1 seconds, proving feasible for robot operation.

\subsection{Robotic Deployment}

\subsubsection{Stage 1: RAPID-SCAN Field Performance}
Field trials with the Clearpath Jackal robot on a six-pipe system (65 feet each) showcased RAPID-SCAN's robust real-time segmentation. It identified defects under varied lighting, debris, and pipe surfaces. The model maintained sub-50ms inference times, ensuring uninterrupted defect detection. Its adaptive processing managed different pipe materials and shapes, providing accurate inputs for VLM analysis. Computational efficiency was crucial for extended operations, maintaining accuracy similar to lab conditions within thermal and power limits.

\subsubsection{Stage 2: VLM Summarization in Field Conditions}
Leveraging RAPID-SCAN's precise real-time segmentation, the VLM component effectively created detailed inspection summaries during field tests. Figure \ref{fig:jackal_experiment} validates our pipeline deployment on the robotic platform. The system deployed its pipeline successfully, generating inspection summaries displayed in the Control Center app for operator analysis. The robust two-stage system consistently detected defects with RAPID-SCAN, and VLM ensured summary quality in challenging environments. It consistently produced actionable summaries matching expert infrastructure assessments, confirming the edge-deployed pipeline's utility for autonomous inspection workflows.

\section{Conclusion}

This paper presents a two-stage pipeline combining the lightweight RAPID-SCAN segmentation model and a fine-tuned vision-language model for automated underground infrastructure inspection. Our approach bridges the gap between defect detection and reporting in robotic inspections. Performance evaluation showed RAPID-SCAN's competitive segmentation (0.834 F1-score, 0.729 mIoU) with high efficiency (0.64M parameters, 0.19 GFLOPS), achieving a 97\% parameter reduction while maintaining accuracy. The fine-tuned VLM produced high-quality summaries validated by experts for better technical terminology and structured inspection principles. Optimizations via QLoRA fine-tuning and TensorRT acceleration allowed real-time edge deployment, with a 3.1-second latency from image capture to summary generation.

Field trials with the Clearpath Jackal robot confirmed the system's robustness in real-world conditions, such as changing lighting and debris. The pipeline created summaries matching expert assessments, proving useful for autonomous inspection. A modular ROS-based setup with Control Center integration improved operator interaction. This research advances embodied AI and human-robot interaction, enabling robots to report in natural language, vital for safety-critical infrastructure. Future work will include broader defect taxonomies, better summarization via VLM architectures, and 3D mapping for accurate spatial context.

 \bibliographystyle{elsarticle-num} 
 \bibliography{references}

\end{document}